\documentclass[runningheads]{llncs}

\usepackage{eccv}

\usepackage{eccvabbrv}

\usepackage{graphicx}
\usepackage{booktabs}
\usepackage{amsmath}
\usepackage{booktabs}
\usepackage[noend]{algpseudocode}
\usepackage{algorithmicx,algorithm}
\usepackage{colortbl}
\usepackage{multirow}
\usepackage{sidecap}
\usepackage{wrapfig}
\usepackage{bbding}

\usepackage[accsupp]{axessibility}  

\usepackage{hyperref}

\usepackage{orcidlink}

\begin{document}

\title{C2C: Component-to-Composition Learning \\ for Zero-Shot Compositional Action Recognition} 

\titlerunning{C2C: Component-to-Composition Learning}

\author{Rongchang~Li\inst{1,2}\orcidlink{0000-0002-5451-4870} \and
Zhenhua~Feng\inst{1,2,3,4}\orcidlink{0000-0002-4485-4249} \and
Tianyang~Xu\inst{1}\orcidlink{0000-0002-9015-3128}\and
Linze~Li\inst{1}\orcidlink{0009-0004-0648-8745} \and
Xiao-Jun~Wu\inst{1}\thanks{Corresponding Author.}\orcidlink{0000-0002-0310-5778}\and
Muhammad~Awais\inst{2,4}\orcidlink{0000-0002-1122-0709} \and
Sara~Atito\inst{2,4}\orcidlink{0000-0002-7576-5791}\and
Josef~Kittler\inst{2,3}\orcidlink{0000-0002-8110-9205}}

\authorrunning{R.~Li et al.}

\institute{School of Artificial Intelligence and Computer Science, Jiangnan University, China \and
Centre for Vision, Speech and Signal Processing (CVSSP), University of Surrey, UK \and
School of Computer Science and Electronic Engineering, University of Surrey, UK \and
Surrey Institute of People-centred AI (SI-PAI), University of Surrey, UK \\
\email{\{li\_rongchang,linze.li\}@stu.jiangnan.edu.cn; \\feng-zhenhua@outlook.com; \{wu\_xiaojun,tianyang\_xu\}@jiangnan.edu.cn;  \{muhammad.awais, sara.atito, j.kittler\}@surrey.ac.uk};
}

\maketitle
\begin{abstract}
Compositional actions consist of dynamic (verbs) and static (objects) concepts. Humans can easily recognize unseen compositions using the learned concepts. For machines, solving such a problem requires a model to recognize unseen actions composed of previously observed verbs and objects, thus requiring so-called compositional generalization ability. To facilitate this research, we propose a novel Zero-Shot Compositional Action Recognition (ZS-CAR) task. For evaluating the task, we construct a new benchmark, Something-composition (Sth-com), based on the widely used Something-Something V2 dataset. We also propose a novel Component-to-Composition (C2C) learning method to solve the new ZS-CAR task. C2C includes an independent component learning module and a composition inference module. Last, we devise an enhanced training strategy to address the challenges of component variations between seen and unseen compositions and to handle the subtle balance between learning seen and unseen actions. The experimental results demonstrate that the proposed framework significantly surpasses the existing compositional generalization methods and sets a new state-of-the-art. The new Sth-com benchmark and code are available at \url{https://github.com/RongchangLi/ZSCAR_C2C}.
\keywords{Zero-shot compositional action recognition \and Action recognition \and Compositional generalization}
\end{abstract}

\section{Introduction}
\label{sec:intro}
Humans can understand basic dynamic and static concepts (\textit{i.e.}, verbs and objects) in seen compositional actions and generalise the concepts to unseen action categories.
For example, by seeing \textit{close a box} and \textit{open a bag}, we can envision \textit{close a bag} even if we have never seen this before.
This compositional generalization capability is a fundamental function of human intelligence, by which we can understand an `infinite' number of complex concepts through assimilating the learned `finite' concepts~\cite{Theory_Syntax, Connectionism}.
Compositional generalization has recently gained attention in the computer vision community, giving rise to the corresponding image analysis task Compositional Zero-Shot Learning (CZSL)~\cite{redwine}.
In video analysis,~\cite{sth_else} proposed the compositional action recognition task.
However, this task only requires recognizing the dynamic component.
It makes the task essentially a regular verb classification task characterized by explicit sample selection bias~\cite{sampling_bias_ori, sampling_bias_icml} regarding verb-object compositions, rather than verifying compositional generalization ability.
Exploring the compositional generalization problem in the video modality is valuable as it presents a challenging video understanding task that requires comprehending both dynamic and static components while also generalising to novel compositions.
Also, it provides a reliable way to develop open-scenario-related applications in autonomous vehicles, robots, etc.

\begin{figure}[t]
  \centering
  \includegraphics[width=.55\linewidth]{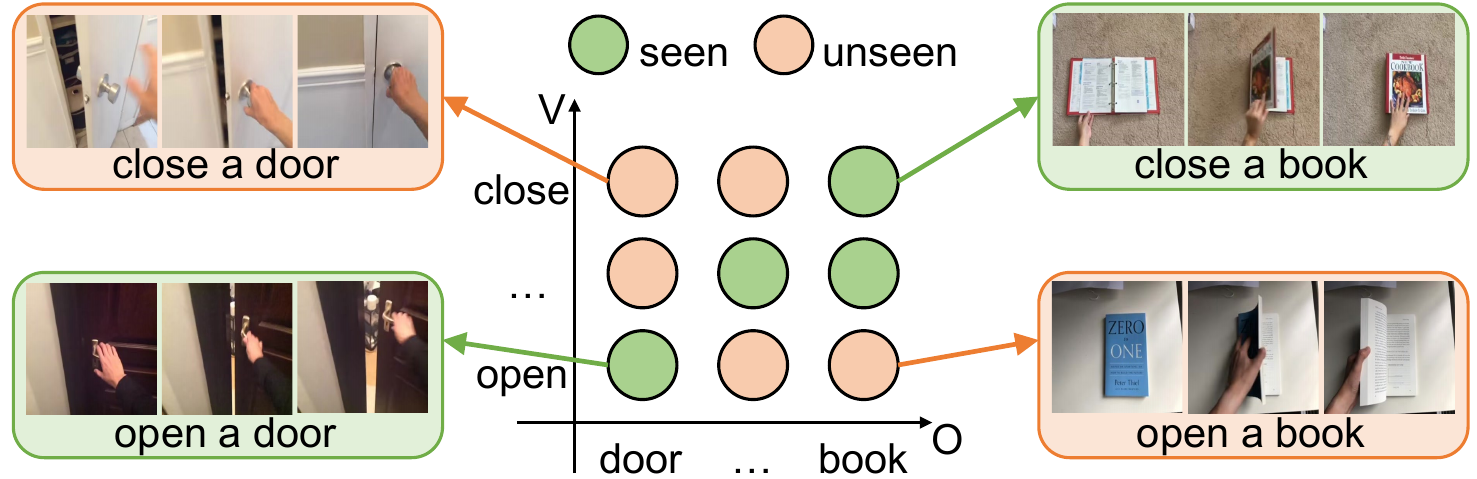}
  \caption{Zero-Shot Compositional Action Recognition (ZS-CAR) requires models to recognize unseen actions composed of verbs and objects observed in seen actions. 
  }
  \label{fig:motivation}
\end{figure}

To facilitate the related research, we propose a novel task, namely \textit{\textbf{Z}ero-\textbf{S}hot \textbf{C}ompositional \textbf{A}ction \textbf{R}ecognition (\textbf{ZS-CAR})}.
We define a compositional \textit{action} as the composition of a dynamic component \textit{verb} (\textit{e.g.}, open, pick up, ...) and a static component \textit{object} (\textit{e.g.}, door, wallet, ...).
As shown in~\cref{fig:motivation}, ZS-CAR aims to enable machines to identify unseen action categories consisting of seen verbs and objects.
The new proposed task poses a significant challenge. 
First, a model has to understand the intricate spatial and temporal information to identify the verb and object components, which is more challenging than the existing action recognition task.
Second, the task naturally contains \textit{component domain variation} and \textit{component compatibility variation}.
The first one is similar to the challenges in domain generalization, caused by the contextuality difference (\textit{e.g.}, transferring knowledge of \textit{book} in \textit{close a book} to recognize \textit{bend a book}).
The second one stems from the action category difference in the training and test domains.
Overly accepting or rejecting verb-object relationships observed during training can detrimentally impact the generalization capability to recognize unseen actions.
Given these challenges, the existing visual compositional generalization solutions cannot handle the ZS-CAR task well, as shown in~\cref{tab:com_baselines}.

To solve ZS-CAR, we present an innovative Component-to-Composition (C2C) method to serve as the baseline.
The pipeline of C2C is shown in~\cref{fig:pipeline}.
C2C first independently learns the dynamic and static components (\textit{i.e.}, verbs and objects).
Then it employs the component consensus to recognize compositions (\textit{i.e.}, actions) from both dynamic and static perspectives, ensuring that both these two components are well learned.
The procedure is motivated by the human process of assembling compositions.
It involves an independent observation of a component, assessing its compatibility with another part, and gradually assembling the units into the entire structure.
Moreover, C2C enables explicitly addressing the \textit{component domain variation} and \textit{component compatibility variation} problems.
Accordingly, we propose an enhanced training strategy that involves minimizing the dependence on spurious information within visual features and balancing the emphasis between fitting seen and imagining unseen actions.

Last, to enable the training of C2C and to evaluate its performance for the ZS-CAR task, we create a new dataset, Something-composition (Sth-com), based on the popular Something-Something V2 (Sth-V2) dataset~\cite{sthv2}.
The experimental results demonstrate that the vanilla or enhanced version of C2C achieves state-of-the-art results.
To summarize, the main contributions of the paper include:
\begin{itemize}
    \item A novel zero-shot compositional action recognition (ZS-CAR) task that investigates compositional generalization in the video modality.
    \item A benchmarking dataset for evaluating ZS-CAR. We elaborate on the challenges in ZS-CAR and demonstrate that the existing visual compositional generalization methods cannot handle such problems well. 
    \item A novel Component-to-Composition (C2C) method as the vanilla solution and an enhanced training strategy to address the identified challenges in ZS-CAR. The experimental results demonstrate that C2C solves these challenges effectively and attains SOTA performances.
\end{itemize}

\section{Related work}
\label{sec:relate}

\textbf{Video action recognition} aims to identify actions in videos.
The main challenge is extracting discriminative action features from videos.
Previous CNN-based methods address this challenge in two primary ways: using 2D CNNs with separate temporal operations~\cite{tsn, tsm, sgm, den, nestnet, tea, msnet, xu2023learning, selfy, tdn}, or jointly using 3D CNNs~\cite{3d, c3d, s3d, r3d, qian2024controllable, i3d, slowfast, x3d} to dig out spatiotemporal features.
Recently, Transformer-based models transform a video into 3D tokens and model the spatiotemporal correlations through attention layers~\cite{timesformer, VTN, vivit, MVitv1, videoswin, uniformer, tubevit}, or adapt the visual image encoder in large vision-language models to the video modality~\cite{stadapter, aim, park2023dual}.


\textbf{Zero-shot action recognition (ZSAR)} aims to recognize action categories not appearing in the training set.
A widely used idea is to measure video features and action embeddings in a common semantic space, hoping this space can generalise to unseen actions.
The semantic embeddings are obtained by side semantic information, \textit{e.g.}, word embeddings~\cite{zsar2020rethinking, zsar2019out, zsar2017zero, zsar2017transductive}, multiple attributes~\cite{zsar2012attribute,zsar2011recognizing}, elaborative rehearsals~\cite{zsar2021ER} and action descriptions~\cite{zsar2022rethinking, zsar2023zero}.
Different from ZSAR, the proposed ZS-CAR aims to explore the visual compositionality of actions and generalize the learned composition knowledge to unseen actions.

\textbf{Compositional action recognition}:
The compositional nature of actions has recently received increasing attention.
Action genome~\cite{act_genome} takes action dynamics as the relationship changes and decomposes actions into spatiotemporal scene graphs.
Something-else~\cite{sth_else} proposes compositional action recognition that only requires recognising verbs in unseen verb-object compositions.
Subsequent studies~\cite{debia_sth_else, yan2022look, ma2022motion, yan2023progressive} follow this setting and focus on extracting robust dynamic information.
However,~\cite{sth_else} essentially is a supervised task with explicit sample selection bias, rather than a compositional generalization problem.

\textbf{Image-based compositional generalization:}
In the image modality, \cite{redwine} first introduces the Compositional Zero-Shot Learning (CZSL) task that requires models trained on seen attribute-object pairs to recognize unseen ones.
A common idea is to find the nearest holistic composition embeddings for input images in a common space~\cite{redwine, compcos, tmn, CGE, protoprop}.
Some studies disentangle visual features to align with component embeddings~\cite{oaid, protoprop, scen, ade, cot, kgsp}.
Recent works propose to adapt the vision-language model~\cite{clip} to CZSL by text prompting methods~\cite{csp,dfsp} and enhancing the interactions between language and visual features~\cite{dfsp}.

\section{Problem Definition and Benchmark}
\label{sec:3-1}

\subsection{The Proposed ZS-CAR Task}

In \textbf{traditional action recognition}, a model predicts the action label defined as either a verb or a verb-object composition.
For example, the Sth-V2 dataset~\cite{sthv2} defines verb-only actions, such as \textit{open sth} and \textit{pick up sth}.
Kinetics-400~\cite{ktc} defines actions as both verb-object compositions and verb-only actions, such as \textit{play basketball}, \textit{draw}, etc.
The recently proposed \textbf{compositional action recognition} task considers the interaction between verbs and objects.
It requires recognising verbs of which the paired objects in the test set differ from those in the training set.
Though introducing action compositionally by this constraint, this task only takes the verb-object relation as a sample selection bias controller, making it not a thorough compositional generalization problem.

In contrast to previous tasks, we introduce the compositional generalization perspective to understand actions, by proposing a novel Zero-Shot Compositional Action Recognition (ZS-CAR) task.
We explicitly define a compositional action as a concept of \textit{verb} and \textit{object}.
To verify the compositional generalization ability, ZS-CAR requires models to recognize unseen action categories composed of verbs and objects observed in seen actions.
Formally, we denote the training set as $D_{\text {train}}=\left\{\left(\mathcal{X}_i, {a}_i\right)\right\}_{i=1}^N$ consisting of video $\mathcal{X}$ labelled as action $a \in {A}_{\text {train}}$, with ${A}_{\text {train}} \subset V\times O=\{(v,o)| v \in V, o \in O\}$, where $V$ and $O$ are the verb and object sets.
The validation and test sets are similarly denoted as $D_{\text{val}}$ and $D_{\text{test}}$, with label sets as $A_{\text{val}}$ and $A_{\text{test}}$.
We follow the generalised zero-shot learning setting~\cite{gzsl, tmn}, \textit{i.e.}, both seen and unseen action categories appear in the validation/test set:
$A_{\text{train}}\cap A_{\text{val/test}} \neq\varnothing$ and $\exists a (a \in A_{\text{val/test}} \oplus a \in A_{\text{train}})$.



\subsection{The Something-composition Benchmark}
\label{sec:4}
To assess the proposed task, we build a benchmark Something-composition (Sth-com), based on the widely used Something-Something V2 (Sth-V2) dataset~\cite{sthv2}.
Sth-V2 has 220,847 videos of humans performing pre-defined compositional actions with everyday objects.
It is widely used by the action recognition community, as its verbs cannot be identified from a single frame.

\begin{table}[!t]
\centering
\caption{Statistics of the proposed Sth-com benchmark.}
\resizebox{.55\textwidth}{!}{%
\begin{tabular}{l|cc|c|c}
\hline
\textbf{} &
  \# Objects &
  \# Verbs &
  \# Verb-Object Actions &
  \# Sample \\ \hline
Train & 248 & 161 & 3451     & 38034 \\ \hline
Val   & 225 & 157 & 733 Seen + 717 Unseen        & 18774 \\
Test  & 233 & 161 & 976 Seen + 956 Unseen        & 22657 \\ \hline
All   & 248 & 161 & 5124 & 79465 \\ \hline
\end{tabular}%
}
\label{tab:sth-com}
\end{table}

\textbf{Annotation}: We use the training and test split of Sth-else~\cite{sth_else}, a subset of Sth-V2, as the starting point. 
Then, we iteratively clean the subsets and ensure that: (1) Each composition (\textit{i.e.}, action) has more than five samples. (2) The components (\textit{i.e.}, verb \& object) present in the test set are seen in the training set.
This yields 248 objects, 161 verbs, 5124 compositions and 79465 video samples.
Next, we select 1/3 of compositions in the training and test sets respectively and interchange 50\% of the selected compositions, thereby introducing seen action categories to the test set.
Subsequently, following the ratio (3:4) in MIT-State~\cite{mitstate}, we divide the test set into the validation and test sets, respectively.
The statistics of Sth-com are summarised in~\cref{tab:sth-com}.


\section{Component-to-Composition Learning}
\begin{figure}[!t]
  \centering
\includegraphics[width=0.95\linewidth]{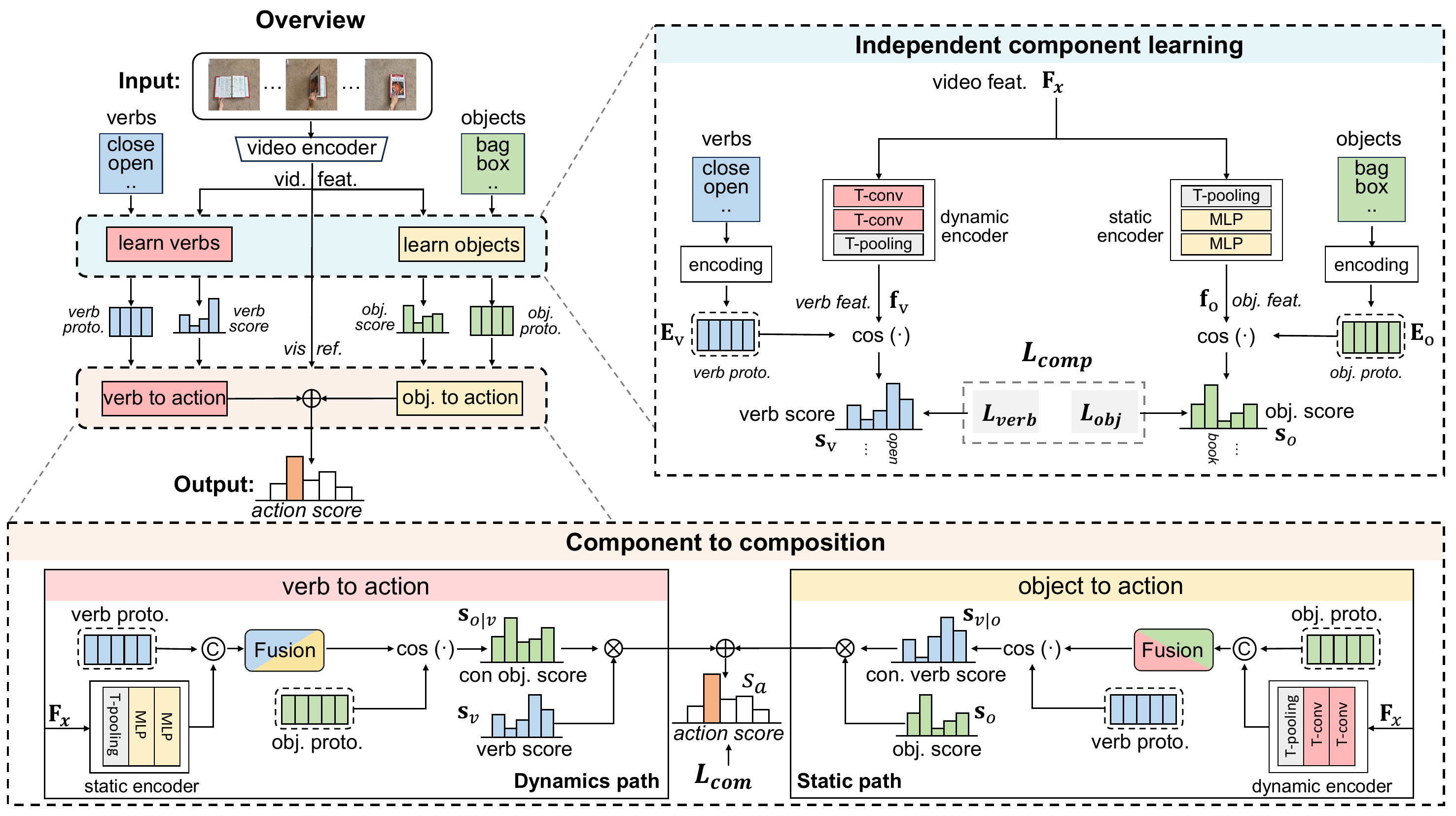}
  \caption{\textbf{The proposed Component-to-Composition (C2C)} learning framework.
    C2C first aligns verb/object prototypes with corresponding visual features to obtain component scores in the \textbf{Independent Component Learning} module.
    Then, actions are inferred through two paths (dynamics and static) in the \textbf{Component to Composition} module.
    In the dynamics path, verb prototypes and visual features are used to compute conditional object scores. Then the independent verb scores and conditional object scores are multiplied to gain the action scores.
    The static path follows a similar procedure.
    The final output is a consensus of the results from both paths.
    }
  \label{fig:pipeline}
\end{figure}

To address the proposed ZS-CAR task, we propose a novel Component-to-Composition (C2C) learning framework, as shown in~\cref{fig:pipeline}, to serve as a vanilla solution.
We simulate the process of human assembly to infer compositions.
We commence with a specific component and scrutinise the compatibility of various categories of another component, leading to a reasonable composition proposal.
These are correspondingly implemented in the proposed \textit{independent component learning} module and \textit{composition to component learning} module.
Importantly, we analyse the challenges in ZS-CAR and devise an \textit{enhanced training strategy} to facilitate generalising to unseen actions better.

\subsection{Independent Component Learning}
\label{sec:4-1}
In this module, our goal is to identify the verb/object score of an input video and acquire their semantic prototypes.
Specifically, we first encode the raw video sequence $\mathcal{X}$ with shape ${T}\times{224}\times{224}\times{3}$ to a general representation $\mathbf{F}_x \in \mathbb{R}^{{T}\times{D}}$, where $T$ and $D$ denote the frame and channel numbers, respectively.
Then, we respectively adopt dynamic and static encoders to extract dynamic and static features, denoted as $\mathbf{f}_v$ and $\mathbf{f}_o \in \mathbb{R}^{C}$, where $C$ is the channel number.
The static encoder consists of a temporal average pooling layer and two MLP layers, while the dynamics encoder contains two temporal convolution layers and a temporal pooling layer.
We encode the component labels as the general representations of verbs and objects, denoted as the component prototypes
$\mathbf{E}_v =\{e_{v,i}\}^{N_v}_{i=1}$ and $\mathbf{E}_o =\{e_{o,i}\}^{N_o}_{i=1}$,
where $N_v$ and $N_o$ are the verb and object numbers, respectively.
The prototypes are learned by being aligned with the visual features $\mathbf{f}_v$ and $\mathbf{f}_o$ using cosine similarity as the gauge.
Denoting the action label of the input video as $a_i=(v_l,o_k)$, the optimization objective is defined as:
\begin{equation}
 \mathcal{L}_{verb}=-\log \frac{\exp \left(cos(\mathbf{f}_v,\mathbf{e}_{v,l})/{\tau}\right)}{\sum_{v_j \in V} \exp \left({cos(\mathbf{f}_v,\mathbf{e}_{v,j})}/{\tau}\right)},
\end{equation}
\begin{equation}
 \mathcal{L}_{obj}=-\log \frac{\exp \left(cos(\mathbf{f}_o,\mathbf{e}_{o,k})/{\tau}\right)}{\sum_{o_j \in O} \exp \left({cos(\mathbf{f}_o,\mathbf{e}_{o,j})}/{\tau}\right)},
\end{equation}
where $\mathbf{e}_{v,l}$ and $\mathbf{e}_{o,k}$ are the corresponding prototypes.
$cos(,)$ is the cosine similarity. 
$\tau$ is the temperature coefficient for better optimization.
We denote $\mathcal{L}_{comp}=\mathcal{L}_{verb}+\mathcal{L}_{obj}$ as the component loss.
By aligning with diverse component features in the training set, the prototypes inherently acquire semantic meanings.
Accordingly, the verb and object scores, $\mathbf{s}_v$ and $\mathbf{s}_o$, are the cosine similarities between the visual features and corresponding component prototypes.

\subsection{From component to composition}
\label{sec:4-2}
In this module, we infer the action (composition) score $\mathbf{s}_a \in \mathbb{R}^{{N_a}}$ using the prototypes and the verbs and objects (components) scores.
We define the conditional scores $\mathbf{S}_{o|v} \in \mathbb{R}^{{N_v}\times{N_o}}$ and $\mathbf{S}_{v|o} \in \mathbb{R}^{{N_o}\times{N_v}}$, that represents the confidence of one component given the visual input and another component as the conditions.
Accordingly, for one action $a_i=(v_l,o_k)$, its confidence score can be calculated by either $s_{a,i}=s_{v,l} \cdot s_{o=o_k|v=v_l}$ or $s_{a,i}=s_{o,k} \cdot s_{v=v_l|o=o_k}$.
The first one relies on the verbs to infer actions, called the dynamics path.
Similarly, the other one is called the static path.
As shown in~\cref{fig:pipeline}, in the dynamics path, we first encode $\mathbf{F}_x$ using a static encoder to provide the visual reference for the object component.
Then we concatenate the output with the verb prototype $\mathbf{e}_{v,l}$ and fuse them to the compound prototype-visual feature $\mathbf{f}_{v-x,l}$, using an MLP layer.
Then we linearly map the cosine similarity $cos(\mathbf{f}_{v-x,l},\mathbf{e}_{o,k})$ to the range of $[0,1]$, obtaining the conditional object score $s_{o=o_k|v=v_l}$.
The static path follows a similar process except for encoding $\mathbf{F}_x$ by the dynamics encoder.
Last, we average the inference results from the two paths to obtain the final inference output.
Taking $s_{a,i}$ as the score of the groudtruth action $a_{i}$, the entire model is optimized using:
\begin{equation}
 \mathcal{L}_{com}=-\log \frac{\exp \left(s_{a,i}/{\tau}\right)}{\sum_{a_{j} \in A_{\text{train}}} \exp \left({s_{a,j}}/{\tau}\right)},
 \label{eq:loss_com}
\end{equation}

\subsection{Enhanced training strategy}
\label{sec:4-3}
\begin{wrapfigure}[13]{r}{0.6\textwidth}
  \centering
  \includegraphics[width=1\linewidth]{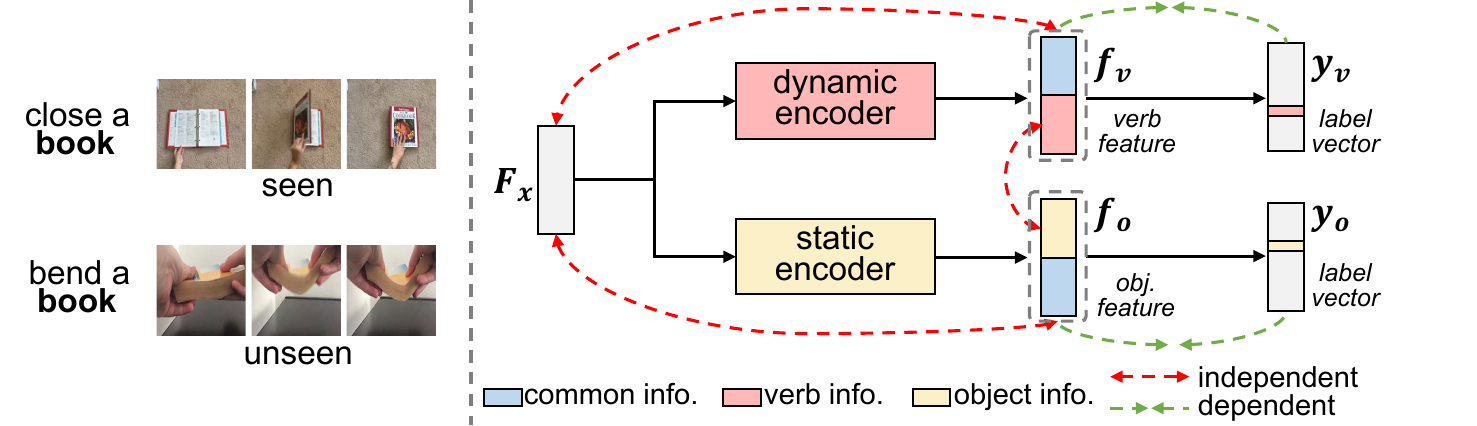}
    \caption{\textbf{Component domain variations}. For unseen actions, an object may exhibit different appearances (left). To deal with this (right), we reduce the spurious information and enlarge the independence between component-specific features. }
  \label{fig:chall_1}
\end{wrapfigure}

\textbf{Risk in generalizing components.} 
Naturally, there is a high probability of exhibiting different component appearances between training and test domains, namely \textit{component domain variation}.
As shown in~\cref{fig:chall_1}, the component `book' exhibits a notable difference in appearance between `close a book' and `bend a book'.
Besides, given the combination of `close' and `book' in the training set, there is a tendency that the model might align features associated with `close' to `book', or vice versa.
These dependencies may confuse the model when encountering unseen action categories.

To cope with it, in the independent component learning module, we posit that the ideal dynamic-focused feature $\mathbf{f}_v$ and static-focused feature $\mathbf{f}_o$ should encompass two types of information: common and specific.
The former denotes features pertinent to both components and the latter denotes the invariant features, that are crucial for identifying the respective components.
Notably, the invariant features of verbs are spurious for identifying objects, and vice versa.
As shown in~\cref{fig:chall_1}, our solution is to minimize the spurious information in $\mathbf{f}_v$ and $\mathbf{f}_o$, and explicitly reduce the mutual associations between their component-specific parts.
To achieve this, we propose the following auxiliary losses:
\begin{equation}
 \mathcal{L}_{sup,verb}= h(\mathbf{f}_x,\mathbf{f}_v)-h(\mathbf{f}_v,\mathbf{y}_v),
\label{eq:verb_hsic}
\end{equation}
\begin{equation}
 \mathcal{L}_{sup,obj}= h(\mathbf{f}_x,\mathbf{f}_o)-h(\mathbf{f}_o,\mathbf{y}_o),
\label{eq:obj_hsic}
\end{equation}
\begin{equation}
 \mathcal{L}_{ind}= \mathcal{L}_{sup,verb}+\mathcal{L}_{sup,obj}+h(\mathbf{f}^{\prime}_v,\mathbf{f}^{\prime}_o),
\end{equation}
where $h(,)$ is the Hilbert-Schmidt Independence Criterion (HSIC)~\cite{hsic} for independence measure.
$h(\mathbf{f}, \mathbf{h}) = 0$ if and only if $\mathbf{f} \perp\!\!\!\!\perp \mathbf{h}$.
$\mathbf{f}_x$ is the temporal pooling result of $\mathbf{F}_x$.
$\mathbf{y}_v$ and $\mathbf{y}_o$ are the one-hot label vectors.
$\mathcal{L}_{sup,verb}$ and $\mathcal{L}_{sup,obj}$ force the model to compress information, but retain sufficient information to identify the components, thus reducing the general spurious information.
$\mathbf{f}^{\prime}_v$ or $\mathbf{f}^{\prime}_o$ is part of $\mathbf{f}_v$ or $\mathbf{f}_o$, defined as $\mathbf{f}^{\prime}_v=\mathbf{f}_v[:\rho C],\mathbf{f}^{\prime}_o=\mathbf{f}_o[:\rho C]$ ($0\leq \rho \leq 1$).
They are seen as component-specific features.
So the term $h(\mathbf{f}^{\prime}_v,\mathbf{f}^{\prime}_o)$ ensures the independence of the invariant part of the verb and object features.

\begin{wrapfigure}[15]{r}{0.52\textwidth}
  \centering
  \includegraphics[width=1\linewidth]{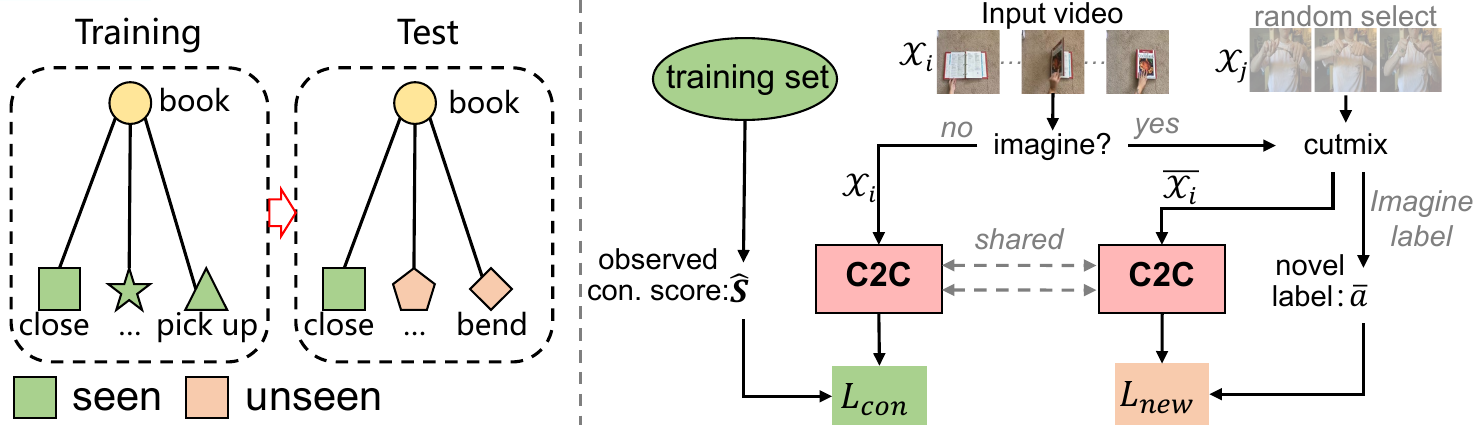}
     \caption{\textbf{Component compatibility variations}. The component relations are different between the training and test sets  (left). To solve this (right), we use the observed conditional score to fit seen relations and encourage the model to imagine unseen actions to avoid being limited to seen relations.
  }
  \label{fig:chall_2}
\end{wrapfigure}

\textbf{Balanced learning of seen \& unseen actions.}
In ZS-CAR, the verb-object connections in the training and test sets differ significantly, as shown in~\cref{fig:chall_2}.
While the seen connections help optimize the modules, an excessive emphasis on them may hinder generalizing to unseen actions.
Our solution is simultaneously enhancing the learning of seen and unseen connections, enabling them to strike an optimal balance.

We first assist the model in \textit{learning seen actions explicitly}.
We calculate the observed conditional probabilities in the training set as the observed conditional scores $\hat{\mathbf{S}}_{v|o}$ and $\hat{\mathbf{S}}_{o|v}$.
Then, we force the conditional scores in a training batch closer to $\hat{\mathbf{S}}_{v|o}$ and $\hat{\mathbf{S}}_{o|v}$, with the objectives:
\begin{equation}
\mathcal{L}_{con,v}=-\frac{1}{N_{v}}\sum_{v_{i} \in {V}} \hat{\mathbf{s}}_{o|v=v_i}\log \mathbf{s}_{o|v=v_i},
\end{equation}
\begin{equation}
\mathcal{L}_{con,o}=-\frac{1}{N_{o}}\sum_{o_{i} \in {O}} \hat{\mathbf{s}}_{v|o=o_i}\log \mathbf{s}_{v|o=o_i},
\end{equation}
where $N_{v}$ \& $N_{o}$ denote the verb \& object numbers, and the condition loss $\mathcal{L}_{con}$ is defined as the sum of $\mathcal{L}_{con,v}$ and $\mathcal{L}_{con,o}$.
$\hat{\mathbf{s}}_{o|v=v_i}$ or $\mathbf{s}_{o|v=v_i}$ denotes the conditional object score vector when verb is $v_i$.
$\hat{\mathbf{s}}_{v|o=o_i}$ and $\mathbf{s}_{v|o=o_i}$ are defined similarly.

Then, inspired by the human ability to create new things by \textit{imagining compositions of known components}, we use CutMix~\cite{yun2019cutmix} to assemble samples in a training batch as the imagined unseen actions, thereby reducing the risk of overemphasizing seen compositions.
Specifically, for an input $\mathcal{X}_i$ with label $a_i=(v_l,o_k)$, we randomly select another sample $\mathcal{X}_j$ with label $a_j=(v_m,o_n)$, and crop it at a random position of random size.
The cropped part replaces the same position of $\mathcal{X}_i$, generating the augmented input $\bar{\mathcal{X}_i}$.
Denoting the cropping ratio as $\lambda$, we modify the previous losses as: $\mathcal {\bar{L}}_{com}=(1-\lambda)\mathcal{L}_{com}^{a_i}+\lambda\mathcal{L}_{com}^{a_j}$ and $ \mathcal {\bar{L}}_{comp}=(1-\lambda)\mathcal{L}_{comp}^{v_l,o_k}+\lambda\mathcal{L}_{comp}^{v_m,o_n}$.
$\mathcal{L}_{com}^{a_i}$ calculates $\mathcal{L}_{com}$ using~\cref{eq:loss_com} with $a_i$ as groudtruth.
$\mathcal{L}_{comp}^{v_l,o_k}$ calculates $\mathcal{L}_{comp}$ with $v_l$ and $o_k$ as groudtruth.
$\mathcal{L}_{com}^{a_j}$ and $\mathcal{L}_{comp}^{v_m,o_n}$ are defined similarly.
Also, $\mathcal {L}_{ind}$ is modified to $\mathcal {\bar{L}}_{ind}$ by replacing $\mathbf{y}_i$ to $(1-\lambda)\mathbf{y}_i+\lambda \mathbf{y}_j$ in~\cref{eq:verb_hsic} and~\cref{eq:obj_hsic}.
Particularly, we introduce new compositions $\bar{a_1}=(v_l,o_n)$ and $\bar{a_2}=(v_m,o_k)$.
$\bar{a_1}$ and $\bar{a_2}$ are not constrained in $A_{\text {test}}$ or $A_{\text {train}}$.
It gives a novel loss $\mathcal {L}_{new}=\mathcal{L}_{com}^{\bar{a_1}}+\mathcal{L}_{com}^{\bar{a_2}}$,
which forces the model to consider all the possible component connections.

A concern is that the new compositions $\bar{a_1}$ or $\bar{a_2}$ may be counterfactual.
However, it only introduces noise to the component-to-composition module, as the component labels still match the visual input.
In this way, we still reserve these counterfactual labels as they can provide regularization effects that avoid overfitting to seen verb-object relations.
Besides, it is also consistent with human imagination which is not only constrained within reasonable compositions.

\begin{table}[t]
\centering
\caption{A comparison with the SOTA methods on Sth-com, using the vision model with word embedding model paradigm.
We report both vanilla (only $\mathcal{L}_{com}$ and $\mathcal{L}_{comp}$) and enhanced (using the advocated enhanced training strategy) versions of C2C.
Since ADE is tailored for transformer architectures, we only adapt it to VideoSwin.}
\resizebox{.85\textwidth}{!}{%
\begin{tabular}{l|cccccc|cccccc}
\hline
              & \multicolumn{6}{c|}{TSM-18}                                                                                                                                 & \multicolumn{6}{c}{VideoSwin-T}                                                                                                                          \\ \hline
Methods       & {\color[HTML]{000000} verb} & obj.          & seen          & unseen        & HM   & AUC  & {\color[HTML]{000000} verb} & obj.          & seen          & unseen        & HM   & AUC  \\ \hline
TMN~\cite{tmn}           & 41.4                        & 35.9          & 31.6          & 28.1          & 22.0          & 7.3           & 33.4   &45.6   &31.0             &30.2               & 23.0            & 7.9              \\
Compcos\cite{compcos}       & 45.8                        & 40.0          & 34.8          & 35.1          & 27.5          & 10.6          & 36.7                        & 50.5          & 34.2          & 37.8          & 27.2          & 10.9          \\
CGE~\cite{CGE}            & 35.5                        & 36.4          & 27.6          & 25.5          & 20.0          & 5.9           &  31.8    & 48.1   &31.2   &30.3               & 23.1              & 7.9             \\
 OADis~\cite{oaid}         & 39.1                        & 40.7          & 33.5          & 32.3          & 25.5          & 9.3           & 41.6      & 48.9              & 38.1              & 38.3              & 30.2             & 12.6             \\
ADE~\cite{ade}           &-                             &-               &-               &-               & -              & -              & 35.3   &49.1  &33.8     & 36.2              & 26.8              & 10.3             \\
CAN~\cite{can}           & 40.0                        & 39.2          & 32.4          & 31.6          & 24.5          & 8.7           &32.6    & 49.5              &33.0               &33.3               & 25.0             & 9.2              \\
\hline
$\textbf{C2C}$ (Vanilla)  & 47.7                        & 42.7          & 38.1          & 39.1          & 30.8          & 13.2          & 47.9                        & 50.9          & 42.1          & 43.6          & 34.9          & 16.4          \\
$\textbf{C2C}$ (Enhanced) & \textbf{50.1}               & \textbf{44.3} & \textbf{39.4} & \textbf{41.0} & \textbf{33.0} & \textbf{14.5} & \textbf{48.8}               & \textbf{52.8} & \textbf{43.4} & \textbf{44.1} & \textbf{36.7} & \textbf{17.4} \\ \hline
\end{tabular}%
}

\label{tab:com_baselines}
\end{table}


\subsection{More details}
\textbf{Network details}.
Following previous practice, we implement the video encoder and component label encoding with two paradigms:

(1) Vision model with word embedding model.
In this paradigm, we adopt an action recognition model (the CNN-based TSM-18~\cite{tsm} or the transformer-based VideoSwin-T~\cite{videoswin}) to extract the general video feature and a word embedding model (fasttext~\cite{fasttext}) to encode the labels.

(2) Vision-language models.
Recently, large pre-trained vision-language models, such as CLIP~\cite{clip}, have shown superior power.
In this paradigm, we adopt the language branch of recent prompt-based compositional generalization methods (Coop~\cite{coop}, CSP~\cite{csp} and SPM~\cite{dfsp}) to encode the labels.
As CLIP are trained using image-text pairs, we apply image-to-video adapters introduced in AIM~\cite{aim} to adapt the CLIP visual encoder (ViT-B/32~\cite{vit_vanilla})  to video modality.

\textbf{Training Objectives}.
When applying the proposed enhanced training strategy, we perform CutMix to generate new compositions with a probability $p$.
If apply CutMix, the final loss is calculated as:
\begin{equation}
 \mathcal {\bar{L}}_{total}=\mathcal {\bar{L}}_{com}+\alpha \mathcal{\bar{L}}_{comp}+\beta \mathcal{\bar{L}}_{ind}+\gamma \mathcal {L}_{new}.
\label{eq:all_cut}
\end{equation}
If not, the final loss is:
\begin{equation}
 \mathcal {L}_{total}=\mathcal{L}_{com}+\alpha \mathcal{L}_{comp}+\beta \mathcal{L}_{ind}+\gamma \mathcal{L}_{con}.
\label{eq:all_nocut}
\end{equation}
As $\mathcal {L}_{new}$ and $\mathcal{L}_{con}$ respectively guide the model focus on unseen and seen actions, we use the same coefficient $\gamma$.

\begin{table}[!t]
\centering
\caption{A comparison with the SOTA CLIP-based methods on Sth-com.}
\resizebox{0.5\columnwidth}{!}{%
\begin{tabular}{l|cccccc}
\hline
methods       & verb          & obj.          & seen          & unseen        & {HM} & {AUC} \\ \hline
Coop~\cite{coop}          & 48.4          & 55.1          & 43.9          & 46.4          & 36.6                                & 18.1                                                      \\
+C2C(vanilla) & 56.2          & 56.9          & 48.3          & 54.6          & 42.7                                & 24.0                                                      \\
+C2C(enhance) & \textbf{56.3} & \textbf{58.9} & \textbf{50.7} & \textbf{56.6} & \textbf{44.5}                       & \textbf{26.0}                                             \\ \hline
CSP~\cite{csp}           & 49.0          & 54.8          & 43.6          & 47.4          & 36.0                                & 18.0                                                      \\
+C2C(vanilla) & 56.3          & 56.6          & 49.0          & 53.4          & 42.5                                & 23.8                                                      \\
+C2C(enhance) & \textbf{57.1} & \textbf{58.6} & \textbf{49.9} & \textbf{56.2} & \textbf{44.3}                       & \textbf{25.5}                                             \\ \hline
SPM~\cite{dfsp}             & 48.9          & 54.5          & 44.0          & 46.3          & 35.5                                & 17.7                                                      \\
+DFM~\cite{dfsp}          & 47.4          & 55.4          & 43.1          & 47.3          & 35.8        & 17.9                              \\
+C2C(vanilla) & 55.7          & 57.3          & 49.3          & 53.4          & 42.2                                & 23.8                                                      \\
+C2C(enhance) & \textbf{56.5} & \textbf{58.4} & \textbf{49.9} & \textbf{55.7} & \textbf{43.8}                       & \textbf{25.2}                                             \\ \hline
\end{tabular}%
}
\label{clip_c2c}
\end{table}

\section{Experiments}
\begin{wrapfigure}[14]{r}{0.5\textwidth}
  \centering
  \includegraphics[height=0.5\linewidth]{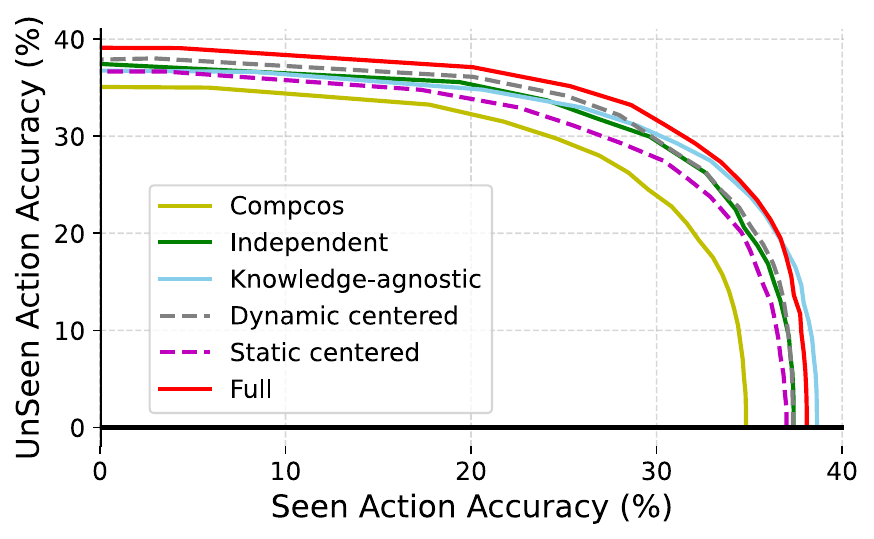}
    \caption{\textbf{Seen-unseen curves}, drawn by gradually forcing the model biased from unseen to seen actions.
}
  \label{fig:diff_hypotheses}
\end{wrapfigure}

\textbf{Evaluation Metrics.}
The metrics include \textbf{Verb}, \textbf{Object}, \textbf{Seen}, \textbf{Unseen}, \textbf{HM} and \textbf{AUC}.
\textbf{Verb} and \textbf{Object} respectively denote the accuracy of verbs and objects.
To gain other metrics, we first add a constant to the seen actions scores.
The constant ranges from negative infinity to positive infinity, making the model shift from completely biased to unseen actions to completely biased to seen actions and yielding the seen-unseen curve shown in~\cref{fig:diff_hypotheses}.
\textbf{Seen} and \textbf{Unseen} are the best accuracy of seen/unseen actions, \textit{i.e.} the intersection points with x/y-axis in~\cref{fig:diff_hypotheses}.
\textbf{HM} is the best harmonic mean with formulation: $HM = 2(Seen \ast Unseen)/(Seen + Unseen)$.
\textbf{AUC} is the area under the curve.

\textbf{Implementation details}.
$\alpha $, $\beta$ $\gamma$ in~\cref{eq:all_cut} and~\cref{eq:all_nocut} are 0.2, 0.1 and 0.1, respectively.
The probability $p$ of conducting CutMix is 0.7.
More coefficient ablations and implementation details are reported in supplementary materials. 

\subsection{Comparisons with State-of-the-Art}
Note that the existing compositional action recognition methods cannot apply to the ZS-CAR task.
To comprehensively demonstrate the proposed C2C, we first adapt the existing CZSL methods to the proposed Sth-com dataset.
We replace their image encoders with video encoders and adopt the same word embedding model, making sure only the compositional generalization parts are different between the adapted methods and our proposed methods.
Besides, we also modify C2C and compare it with the existing SOTA methods on the image-based CZSL task using the challenging C-GQA~\cite{CGE} dataset.

\begin{table}[!t]
\centering
\caption{A comparison with the SOTA methods on C-GQA.}
\resizebox{0.42\columnwidth}{!}{%
\centering
\begin{tabular}{l|c|cccc }
\hline
      &  multi & seen & unseen & {HM}   & {AUC} \\ \hline
CGE~\cite{CGE} &$\times$& 38.0 & 17.1   & 18.5 & 5.4 \\
OADis~\cite{oaid}   &$\times$& 38.3 & 19.8   & 20.1 & 7.0 \\
CoT~\cite{cot}  &$\surd$& 39.2 & 22.7   & 22.1 & 7.4 \\ \hline
\textbf{C2C}    &$\times$& 39.1 & 23.0   & 23.0 & 7.7 \\
\textbf{C2C}   &$\surd$& \textbf{39.4} & \textbf{23.6}   & \textbf{23.3} & \textbf{8.2} \\ \hline
\end{tabular}%
}
\label{tab:cgqa}
\end{table}

\begin{table}[!t]
\centering
\caption{Results of different hypotheses inferring compositions from components.}
\resizebox{0.55\columnwidth}{!}{%
\begin{tabular}{l|cccccc}
\hline
                   & verb             & obj.            & seen            & unseen             & {HM}   & {AUC}  \\ \hline
Compcos\cite{compcos}            & 45.8          & 40.0          & 34.8          & 35.1          & 27.5          & 10.6          \\ \hline
Independent        & 47.9          & 42.2          & 37.4          & 37.4          & 29.8          & 12.2          \\
Knowledge-agnostic & \textbf{48.4} & \textbf{43.1} & \textbf{38.7} & 36.8          & 30.2          & 12.6          \\ \hline
Dynamic centered     & 46.7          & 42.5          & 37.4          & 38.0          & 29.9          & 12.5          \\
Static centered    & 46.6          & 41.1          & 37.0          & 36.7          & 28.9          & 11.7          \\ 
Full        & 47.7 & 42.7 & 38.1 & \textbf{39.1} & \textbf{30.8} & \textbf{13.2} \\ \hline
\end{tabular}%
}
\label{tab:ab_infer}
\end{table}

\textbf{Comparisons on Sth-com.}
We first compare our method with the adapted CZSL methods of the vision model with the word embedding paradigm, using both CNN-based and transformer-based video encoders.
The results are reported in~\cref{tab:com_baselines}. The vanilla version of C2C uses two basic losses. The enhanced version uses the proposed enhanced training strategy.

\cref{tab:com_baselines} demonstrate that our method outperforms the existing methods by significant margins across all the metrics on ZS-CAR.
Furthermore, the proposed enhanced training strategy provides consistent performance gains across different video encoders.
\cref{tab:com_baselines} shows that the recently proposed methods based on Compcos are inferior to Compcos.
Second, except for OADis, using the recently proposed VideoSwin-T backbone does not yield significant improvements.
The first observation may stem from failing to generalise the embedding relations, not disentangling spatial and temporal information well, and not applying the conditional component relations in actions.
Besides, jointly learning the compositions may lead to unbalanced learning of the components.
These deficiencies degrade the effectiveness of video encoders, leading directly to the second observation.
\cref{tab:com_baselines} also compares two video encoders.
Though VideoSwin is an advanced model, its verb accuracy is consistently inferior or similar to TSM across all methods.
Conversely, it outperforms TSM in object accuracy.
This implies that our ZS-CAR task offers a new avenue for assessing biases regarding dynamic and static patterns in action recognition models.

Then we apply our C2C method to recently proposed CLIP-based compositional zero-shot learning methods.
The results are reported in~\cref{clip_c2c}.
Coop, CSP and SPM are three text prompting methods.
We can see that the proposed C2C method can improve their performance significantly on the ZS-CAR task.
DFM~\cite{dfsp} was proposed along with SPM. 
It enables cross-modal interactions between language and visual features for better composition prediction.
However, the improvements are slight on the ZS-CAR task.
Compared to it, our C2C method provides more significant improvements.

\textbf{Comparision on C-GQA}.
To further evaluate the proposed C2C method, we modify it for the image-based compositional zero-shot learning task by deleting all the temporal-related operations.
We report the results on the recently proposed challenging benchmark C-GQA in~\cref{tab:cgqa}, with fine-tuned ViT-B~\cite{vit_vanilla} as the backbone.
In the table, the term `multi' denotes the use of multi-level backbone outputs. When using multi-level outputs in C2C, we merely average them, much simpler than CoT.
Though designed for video analysis, the proposed C2C exhibits superior performance over existing SOTA methods.
Like CoT~\cite{cot}, we also try to utilize multi-level backbone outputs.
But compared to the specially designed module in CoT, we merely average the multi-level outputs.
This demonstrates that our C2C method can be a general solution to the compositional generalization problem.

\subsection{Ablation Study}
\textbf{Different composition generation methods.}
We first compare different methods that can be used to generate compositions from verb and object components, including:
(1) \textbf{Independent} supposes components are completely independent, so the inference is straightforward: $s_{c,i}=s_{v,i} \cdot s_{o,i}$.
(2) \textbf{Knowledge-agnostic} assumes components are not completely independent, but the component relations are agnostic:$s_{c,i}=s_{v,i} \cdot {1} + s_{o,i} \cdot {1}$.
(3) \textbf{One path inference} considers the compatibility between compositions but only relies on either inference path, \textit{i.e.}, the dynamics or the static path result (\cref{fig:pipeline}).
(4) \textbf{Full} is the final scheme, integrating both paths.

For a fair comparison, the experiments only use the base losses $ \mathcal{L}_{com}$ and $ \mathcal{L}_{comp}$.
The results are reported in~\cref{tab:ab_infer}.
We use Compcos as the baseline result.
The simplest inference method, \textit{i.e.}, \textit{Independent}, has surpassed Compcos on all the metrics.
When using the knowledge-agnostic hypothesis, the seen accuracy and holistic metrics are better.
But its unseen accuracy is lower.
By considering the component relations, utilizing both inference paths substantially improves the unseen accuracy, yielding the best results for both HM and AUC. 
The results suggest that our proposed inference approach achieves a more balanced performance between seen and unseen actions.
\cref{fig:diff_hypotheses} reports the seen-unseen curves, implying that the proposed method is more robust.

\begin{table}[!t]
\centering
\caption{Effectiveness of the elements in the enhanced training strategy.}
\resizebox{0.4\columnwidth}{!}{%
\begin{tabular}{cccccc|cc }
\hline
$\mathcal{L}_{com}$      & $\mathcal{L}_{comp}$      & $\mathcal{L}_{ind}$  & CutMix & $\mathcal{L}_{con}$  &$\mathcal{L}_{new}$     & HM   & AUC  \\ \hline
$\surd$                    &                           &                     &           &      &                  & 29.8          & 12.5          \\
$\surd$                    & $\surd$                   &                     &           &      &                 & 30.8          & 13.2          \\ \hline
$\surd$                    & $\surd$                      & $\surd$               &           &    &                   & 31.4          & 13.6          \\ \hline

$\surd$                    & $\surd$                    & $\surd$              &$\surd$   &     &                & 31.5 & 13.9\\ 
$\surd$                    & $\surd$                    & $\surd$              &$\surd$   &$\surd$     &                & 32.1 & 14.0\\ 
$\surd$                    & $\surd$                    & $\surd$              &$\surd$      & $\surd$       &$\surd$           & \textbf{33.0} & \textbf{14.5} \\ \hline

\end{tabular}%
}
\label{tab:ab_learn}
\end{table}

\begin{table}[!t]
\centering
\caption{The accuracy (\%) on actions containing verbs that cause deformations.
}
\resizebox{.5\columnwidth}{!}{%
\begin{tabular}{c|cc|cc|cc}
\hline
\multirow{2}{*}{methods} & \multicolumn{2}{c|}{verb} & \multicolumn{2}{c|}{object} & \multicolumn{2}{c}{composition} \\ \cline{2-7} 
                              & seen & unseen & seen & unseen & seen & unseen \\ \hline
Vanilla                       & 70.1 & 60.2   & \textbf{68.7} & 34.8   & 56.9 & 27.5   \\
Vanilla + $\mathcal{L}_{ind}$ & \textbf{71.8} & \textbf{65.2}   & 68.2 & \textbf{45.1}   & \textbf{58.0} & \textbf{36.7}   \\ \hline
\end{tabular}%
}
\label{tab:hsic}
\end{table}

\begin{table}[!t]
\centering
\caption{Unbiased accuracy (\%) of seen and unseen actions and the corresponding harmonic mean (HM).
}
\resizebox{.7\columnwidth}{!}{%
\begin{tabular}{c|cc|c}
\hline
methods                & seen          & unseen          & HM          \\ \hline
baseline                        & 36.8          & 17.2            & 23.5        \\ \hline
+$\mathcal{L}_{con}$            & 36.5(-0.3)    & 16.4(-0.8)      & 22.6(-0.9) \\
+cutmix                         & 35.7(-1.1)    & 17.7(+0.5)      & 23.6(+0.1) \\ \hline
+cutmix+$\mathcal{L}_{con}$     & 36.1(+0.4)    & 17.6(-0.1)      & 23.7(+0.1) \\
+cutmix+$\mathcal{L}_{new}$     & 37.0(+1.3)    & 18.2(+0.5)      & 24.4(+0.8) \\ \hline
+cutmix+$\mathcal{L}_{con}$+$\mathcal{L}_{new}$  & \textbf{37.5}(+0.5/+1.4)  & \textbf{18.7}(+0.5/+1.1)     & \textbf{24.9}(+0.5/+1.2)\\ \hline
\end{tabular}%
}
\label{tab:balance_learn}
\end{table}

\textbf{The enhanced training strategy}.
In \cref{tab:ab_learn}, we demonstrate the effectiveness of our enhanced training strategy.
It shows that the results exhibit consistent improvements as gradually applying the proposed strategy.

To further validate the proposed method, we verify the merits of the independent loss $L_{ind}$ on actions containing verbs that can cause object deformations (such as \textit{squeeze} and \textit{twist}).
The results are shown in~\cref{tab:hsic}.
They show that the unseen accuracy improves significantly after adding $L_{ind}$.
However, the object accuracy of seen actions slightly decreases.
This might be because spurious features are useful for recognizing seen components.

Next, we validate the proposed balanced learning of actions.
With the vanilla version with $L_{ind}$ as the baseline, we report the accuracy of all seen and unseen actions and HM without adding bias in~\cref{tab:balance_learn}.
It shows that adding only $L_{con}$ leads to performance degradation, but the degradation in seen accuracy is minor.
Similarly, adding just CutMix results in a slight improvement, but with an obvious degradation in seen accuracy.
When using CutMix, we add either $L_{con}$ or $L_{new}$.
The results indicate that $L_{new}$ yields a universal improvement, while $L_{con}$ contributes positively to seen accuracy.
Last, when $L_{new}$ and $L_{con}$ are integrated, they both bring improvements on all the metrics, and the gains are more pronounced than adding either one alone.
This indicates that our training strategy enables the model to achieve a better balance between seen and unseen actions.

\subsection{Qualitative analysis}
\begin{figure}[!t]
\centering
\begin{minipage}[t]{0.48\textwidth}
\centering
\includegraphics[width=0.9\linewidth]{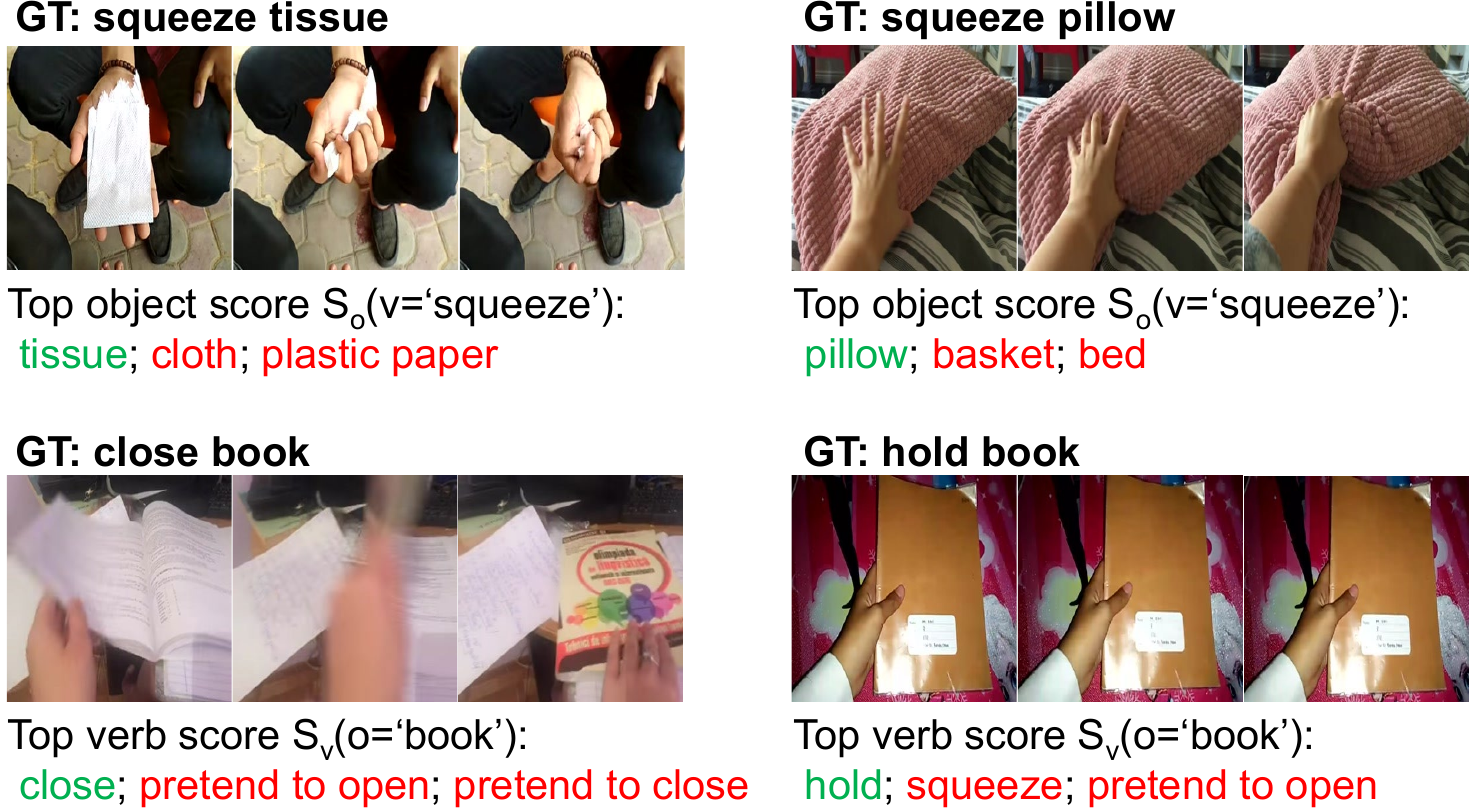}
\caption{The top-3 compatible components with the same verb or object components.}
\label{fig:con_score}
\end{minipage}
\begin{minipage}[t]{0.03\textwidth}
\end{minipage}
\begin{minipage}[t]{0.48\textwidth}
\centering
\includegraphics[width=0.75\linewidth]{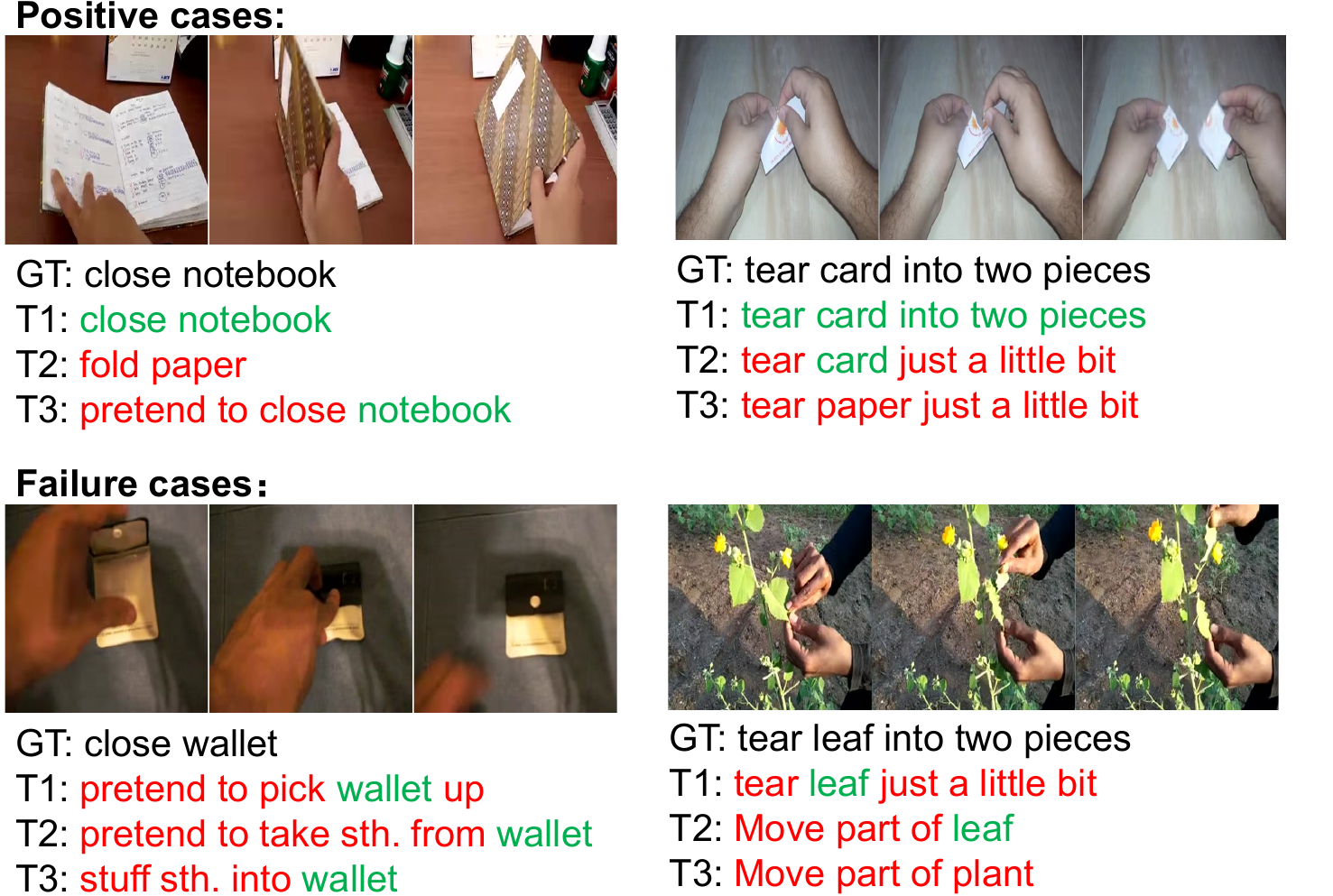}
\caption{Examples of successes and failures of our methods.}
\label{fig:suc_fail_sample}
\end{minipage}
\end{figure}

In~\cref{fig:con_score}, we show the top three compatible components for different samples, while keeping another component the same.
When the verb is \textit{squeeze}, for the first sample, the second and third compatible objects are \textit{cloth} and \textit{paper}, which are close to the visual contents and compatible with \textit{squeeze}.
For the second sample, the model identifies different objects.
The second object \textit{basket} is not included in the visual content but is compatible with \textit{squeeze}.
The third object \textit{bed} cannot be combined with \textit{squeeze} but is visually close to the sample.
When the fixed component is an object, the situation is similar.
It demonstrates that the proposed model can identify component compatibility according to both visual contents and component attributions.

In~\cref{fig:suc_fail_sample}, we visualize the top three recognized actions of some successful and failed samples.
In the top row, our model accurately predicts the given actions.
Besides, the second and third predictions relate to the visual content, such as \textit{fold paper} for the first sample and \textit{tear paper just a little bit} for the second sample.
The bottom row shows some failures.
Though wrong, the predictions still reflect the video content, such as \textit{move part of plant} and \textit{tear leaf into two pieces} for the second example.
However, the failures also demonstrate some limitations of the proposed method. 
For example, it struggles with the task of distinguishing objects or verbs of similar visual appearances.

\section{Conclusion}
This paper introduces a novel task, Zero-Shot Compositional Action Recognition (ZS-CAR), that investigates the compositional generalization in the video modality.
A benchmarking dataset Sth-com and a baseline method C2C are proposed along with the novel task.
Besides, an enhanced training strategy is presented to address two inherent challenges in ZS-CAR: component domain variations and component compatibility variations.
The approach demonstrates superior performance compared to existing compositional generalization solutions in both image- and video-based zero-shot compositional learning settings.

\section*{Acknowledgements}
This work is supported in part by the National Key Research and Development Program of China (2023YFF1105102, 2023YFF1105105), the National Natural Science Foundation of China (62020106012, 62332008, 62106089, U1836218, 62336004), the 111 Project of Ministry of Education of China (B12018), the Postgraduate Research \& Practice Innovation Program of Jiangsu Province (KYCX22\\\_2307), and the UK EPSRC (EP/V002856/1, EP/T022205/1).

%
%
\bibliographystyle{splncs04}

\begin{thebibliography}{10}
\providecommand{\url}[1]{\texttt{#1}}
\providecommand{\urlprefix}{URL }
\providecommand{\doi}[1]{https://doi.org/#1}

\bibitem{vivit}
Arnab, A., Dehghani, M., Heigold, G., Sun, C., Lu{\v{c}}i{\'c}, M., Schmid, C.: Vivit: A video vision transformer. In: Proceedings of the IEEE/CVF International Conference on Computer Vision. pp. 6836--6846 (2021)

\bibitem{timesformer}
Bertasius, G., Wang, H., Torresani, L.: Is space-time attention all you need for video understanding? In: International Conference on Machine Learning. pp. 813--824 (2021)

\bibitem{fasttext}
Bojanowski, P., Grave, E., Joulin, A., Mikolov, T.: Enriching word vectors with subword information. Transactions of the association for computational linguistics  \textbf{5},  135--146 (2017)

\bibitem{zsar2020rethinking}
Brattoli, B., Tighe, J., Zhdanov, F., Perona, P., Chalupka, K.: Rethinking zero-shot video classification: End-to-end training for realistic applications. In: Proceedings of the IEEE/CVF Conference on Computer Vision and Pattern Recognition. pp. 4613--4623 (2020)

\bibitem{i3d}
Carreira, J., Zisserman, A.: Quo vadis, action recognition? a new model and the kinetics dataset. In: Proceedings of the IEEE/CVF Conference on Computer Vision and Pattern Recognition. pp. 6299--6308 (2017)

\bibitem{gzsl}
Chao, W.L., Changpinyo, S., Gong, B., Sha, F.: An empirical study and analysis of generalized zero-shot learning for object recognition in the wild. In: Proceedings of the European Conference on Computer Vision. pp. 52--68 (2016)

\bibitem{zsar2021ER}
Chen, S., Huang, D.: Elaborative rehearsal for zero-shot action recognition. In: Proceedings of the IEEE/CVF International Conference on Computer Vision. pp. 13638--13647 (2021)

\bibitem{Theory_Syntax}
Chomsky, N.: Aspects of the Theory of Syntax. No.~11, MIT press (2014)

\bibitem{zsar2023zero}
Doshi, K., Yilmaz, Y.: Zero-shot action recognition with transformer-based video semantic embedding. In: Proceedings of the IEEE/CVF Conference on Computer Vision and Pattern Recognition. pp. 4858--4867 (2023)

\bibitem{vit_vanilla}
Dosovitskiy, A., Beyer, L., Kolesnikov, A., Weissenborn, D., Zhai, X., Unterthiner, T., Dehghani, M., Minderer, M., Heigold, G., Gelly, S., et~al.: An image is worth 16x16 words: Transformers for image recognition at scale. In: International Conference on Learning Representations (2020)

\bibitem{MVitv1}
Fan, H., Xiong, B., Mangalam, K., Li, Y., Yan, Z., Malik, J., Feichtenhofer, C.: Multiscale vision transformers. In: Proceedings of the IEEE/CVF International Conference on Computer Vision. pp. 6824--6835 (2021)

\bibitem{x3d}
Feichtenhofer, C.: X3d: Expanding architectures for efficient video recognition. In: Proceedings of the IEEE/CVF Conference on Computer Vision and Pattern Recognition. pp. 203--213 (2020)

\bibitem{slowfast}
Feichtenhofer, C., Fan, H., Malik, J., He, K.: Slowfast networks for video recognition. In: Proceedings of the IEEE/CVF International Conference on Computer Vision. pp. 6202--6211 (2019)

\bibitem{Connectionism}
Fodor, J.A., Pylyshyn, Z.W.: Connectionism and cognitive architecture: A critical analysis. Cognition  \textbf{28}(1-2),  3--71 (1988)

\bibitem{zsar2012attribute}
Fu, Y., Hospedales, T.M., Xiang, T., Gong, S.: Attribute learning for understanding unstructured social activity. In: Proceedings of the European Conference on Computer Vision. pp. 530--543 (2012)

\bibitem{sthv2}
Goyal, R., Ebrahimi~Kahou, S., Michalski, V., Materzynska, J., Westphal, S., Kim, H., Haenel, V., Fruend, I., Yianilos, P., Mueller-Freitag, M., et~al.: The" something something" video database for learning and evaluating visual common sense. In: Proceedings of the IEEE/CVF International Conference on Computer Vision. pp. 5842--5850 (2017)

\bibitem{hsic}
Gretton, A., Bousquet, O., Smola, A., Sch{\"o}lkopf, B.: Measuring statistical dependence with hilbert-schmidt norms. In: International Conference on Algorithmic Learning Theory. pp. 63--77 (2005)

\bibitem{ade}
Hao, S., Han, K., Wong, K.Y.K.: Learning attention as disentangler for compositional zero-shot learning. In: Proceedings of the IEEE/CVF Conference on Computer Vision and Pattern Recognition. pp. 15315--15324 (2023)

\bibitem{r3d}
Hara, K., Kataoka, H., Satoh, Y.: Can spatiotemporal 3d cnns retrace the history of 2d cnns and imagenet? In: Proceedings of the IEEE/CVF Conference on Computer Vision and Pattern Recognition. pp. 6546--6555 (2018)

\bibitem{sampling_bias_ori}
Heckman, J.J.: Sample selection bias as a specification error. Econometrica: Journal of the econometric society pp. 153--161 (1979)

\bibitem{mitstate}
Isola, P., Lim, J.J., Adelson, E.H.: Discovering states and transformations in image collections. In: Proceedings of the IEEE/CVF Conference on Computer Vision and Pattern Recognition. pp. 1383--1391 (2015)

\bibitem{act_genome}
Ji, J., Krishna, R., Fei-Fei, L., Niebles, J.C.: Action genome: Actions as compositions of spatio-temporal scene graphs. In: Proceedings of the IEEE/CVF Conference on Computer Vision and Pattern Recognition. pp. 10236--10247 (2020)

\bibitem{3d}
Ji, S., Xu, W., Yang, M., Yu, K.: 3d convolutional neural networks for human action recognition. IEEE Transactions on Pattern Analysis and Machine Intelligence  \textbf{35}(1),  221--231 (2012)

\bibitem{kgsp}
Karthik, S., Mancini, M., Akata, Z.: Kg-sp: Knowledge guided simple primitives for open world compositional zero-shot learning. In: Proceedings of the IEEE/CVF Conference on Computer Vision and Pattern Recognition. pp. 9336--9345 (2022)

\bibitem{ktc}
Kay, W., Carreira, J., Simonyan, K., Zhang, B., Hillier, C., Vijayanarasimhan, S., Viola, F., Green, T., Back, T., Natsev, P., et~al.: The kinetics human action video dataset. arXiv preprint arXiv:1705.06950  (2017)

\bibitem{cot}
Kim, H., Lee, J., Park, S., Sohn, K.: Hierarchical visual primitive experts for compositional zero-shot learning. In: Proceedings of the IEEE/CVF International Conference on Computer Vision. pp. 5675--5685 (2023)

\bibitem{msnet}
Kwon, H., Kim, M., Kwak, S., Cho, M.: Motionsqueeze: Neural motion feature learning for video understanding. In: Proceedings of the European Conference on Computer Vision. pp. 345--362 (2020)

\bibitem{selfy}
Kwon, H., Kim, M., Kwak, S., Cho, M.: Learning self-similarity in space and time as generalized motion for video action recognition. In: Proceedings of the IEEE/CVF International Conference on Computer Vision. pp. 13065--13075 (2021)

\bibitem{uniformer}
Li, K., Wang, Y., Gao, P., Song, G., Liu, Y., Li, H., Qiao, Y.: Uniformer: Unified transformer for efficient spatiotemporal representation learning. arXiv preprint arXiv:2201.04676  (2022)

\bibitem{den}
Li, R.C., Wu, X.J., Wu, C., Xu, T.Y., Kittler, J.: Dynamic information enhancement for video classification. Image and Vision Computing  \textbf{114},  104244 (2021)

\bibitem{sgm}
Li, R., Wu, X.J., Xu, T.: Video is graph: Structured graph module for video action recognition. arXiv preprint arXiv:2110.05904  (2021)

\bibitem{nestnet}
Li, R., Xu, T., Wu, X.J., Shen, Z., Kittler, J.: Perceiving actions via temporal video frame pairs. ACM Transactions on Intelligent Systems and Technology  \textbf{15}(3),  1--20 (2024)

\bibitem{scen}
Li, X., Yang, X., Wei, K., Deng, C., Yang, M.: Siamese contrastive embedding network for compositional zero-shot learning. In: Proceedings of the IEEE/CVF Conference on Computer Vision and Pattern Recognition. pp. 9326--9335 (2022)

\bibitem{tea}
Li, Y., Ji, B., Shi, X., Zhang, J., Kang, B., Wang, L.: Tea: Temporal excitation and aggregation for action recognition. In: Proceedings of the IEEE/CVF Conference on Computer Vision and Pattern Recognition. pp. 909--918 (2020)

\bibitem{tsm}
Lin, J., Gan, C., Han, S.: Tsm: Temporal shift module for efficient video understanding. In: Proceedings of the IEEE/CVF International Conference on Computer Vision. pp. 7083--7093 (2019)

\bibitem{zsar2011recognizing}
Liu, J., Kuipers, B., Savarese, S.: Recognizing human actions by attributes. In: Proceedings of the IEEE/CVF Conference on Computer Vision and Pattern Recognition. pp. 3337--3344 (2011)

\bibitem{videoswin}
Liu, Z., Ning, J., Cao, Y., Wei, Y., Zhang, Z., Lin, S., Hu, H.: Video swin transformer. In: Proceedings of the IEEE/CVF Conference on Computer Vision and Pattern Recognition. pp. 3202--3211 (2022)

\bibitem{dfsp}
Lu, X., Guo, S., Liu, Z., Guo, J.: Decomposed soft prompt guided fusion enhancing for compositional zero-shot learning. In: Proceedings of the IEEE/CVF Conference on Computer Vision and Pattern Recognition. pp. 23560--23569 (2023)

\bibitem{ma2022motion}
Ma, L., Zheng, Y., Zhang, Z., Yao, Y., Fan, X., Ye, Q.: Motion stimulation for compositional action recognition. IEEE Transactions on Circuits and Systems for Video Technology  (2022)

\bibitem{compcos}
Mancini, M., Naeem, M.F., Xian, Y., Akata, Z.: Open world compositional zero-shot learning. In: Proceedings of the IEEE/CVF Conference on Computer Vision and Pattern Recognition. pp. 5222--5230 (2021)

\bibitem{zsar2019out}
Mandal, D., Narayan, S., Dwivedi, S.K., Gupta, V., Ahmed, S., Khan, F.S., Shao, L.: Out-of-distribution detection for generalized zero-shot action recognition. In: Proceedings of the IEEE/CVF Conference on Computer Vision and Pattern Recognition. pp. 9985--9993 (2019)

\bibitem{sth_else}
Materzynska, J., Xiao, T., Herzig, R., Xu, H., Wang, X., Darrell, T.: Something-else: Compositional action recognition with spatial-temporal interaction networks. In: Proceedings of the IEEE/CVF Conference on Computer Vision and Pattern Recognition. pp. 1049--1059 (2020)

\bibitem{redwine}
Misra, I., Gupta, A., Hebert, M.: From red wine to red tomato: Composition with context. In: Proceedings of the IEEE/CVF Conference on Computer Vision and Pattern Recognition. pp. 1792--1801 (2017)

\bibitem{CGE}
Naeem, M.F., Xian, Y., Tombari, F., Akata, Z.: Learning graph embeddings for compositional zero-shot learning. In: Proceedings of the IEEE/CVF Conference on Computer Vision and Pattern Recognition. pp. 953--962 (2021)

\bibitem{csp}
Nayak, N.V., Yu, P., Bach, S.: Learning to compose soft prompts for compositional zero-shot learning. In: The Eleventh International Conference on Learning Representations (2022)

\bibitem{VTN}
Neimark, D., Bar, O., Zohar, M., Asselmann, D.: Video transformer network. In: Proceedings of the IEEE/CVF International Conference on Computer Vision. pp. 3163--3172 (2021)

\bibitem{stadapter}
Pan, J., Lin, Z., Zhu, X., Shao, J., Li, H.: St-adapter: Parameter-efficient image-to-video transfer learning. Advances in Neural Information Processing Systems  \textbf{35},  26462--26477 (2022)

\bibitem{park2023dual}
Park, J., Lee, J., Sohn, K.: Dual-path adaptation from image to video transformers. In: Proceedings of the IEEE/CVF Conference on Computer Vision and Pattern Recognition. pp. 2203--2213 (2023)

\bibitem{tubevit}
Piergiovanni, A., Kuo, W., Angelova, A.: Rethinking video vits: Sparse video tubes for joint image and video learning. In: Proceedings of the IEEE/CVF Conference on Computer Vision and Pattern Recognition. pp. 2214--2224 (2023)

\bibitem{tmn}
Purushwalkam, S., Nickel, M., Gupta, A., Ranzato, M.: Task-driven modular networks for zero-shot compositional learning. In: Proceedings of the IEEE/CVF International Conference on Computer Vision. pp. 3593--3602 (2019)

\bibitem{qian2024controllable}
Qian, R., Lin, W., See, J., Li, D.: Controllable augmentations for video representation learning. Visual Intelligence  \textbf{2}(1), ~1 (2024)

\bibitem{zsar2022rethinking}
Qian, Y., Yu, L., Liu, W., Hauptmann, A.G.: Rethinking zero-shot action recognition: Learning from latent atomic actions. In: Proceedings of the European Conference on Computer Vision. pp. 104--120 (2022)

\bibitem{clip}
Radford, A., Kim, J.W., Hallacy, C., Ramesh, A., Goh, G., Agarwal, S., Sastry, G., Askell, A., Mishkin, P., Clark, J., et~al.: Learning transferable visual models from natural language supervision. In: International conference on machine learning. pp. 8748--8763. PMLR (2021)

\bibitem{protoprop}
Ruis, F., Burghouts, G., Bucur, D.: Independent prototype propagation for zero-shot compositionality. Advances in Neural Information Processing Systems  \textbf{34},  10641--10653 (2021)

\bibitem{oaid}
Saini, N., Pham, K., Shrivastava, A.: Disentangling visual embeddings for attributes and objects. In: Proceedings of the IEEE/CVF Conference on Computer Vision and Pattern Recognition. pp. 13658--13667 (2022)

\bibitem{debia_sth_else}
Sun, P., Wu, B., Li, X., Li, W., Duan, L., Gan, C.: Counterfactual debiasing inference for compositional action recognition. In: Proceedings of the ACM International Conference on Multimedia. pp. 3220--3228 (2021)

\bibitem{c3d}
Tran, D., Bourdev, L., Fergus, R., Torresani, L., Paluri, M.: Learning spatiotemporal features with 3d convolutional networks. In: Proceedings of the IEEE/CVF International Conference on Computer Vision. pp. 4489--4497 (2015)

\bibitem{tdn}
Wang, L., Tong, Z., Ji, B., Wu, G.: Tdn: Temporal difference networks for efficient action recognition. In: Proceedings of the IEEE/CVF Conference on Computer Vision and Pattern Recognition. pp. 1895--1904 (2021)

\bibitem{tsn}
Wang, L., Xiong, Y., Wang, Z., Qiao, Y., Lin, D., Tang, X., Van~Gool, L.: Temporal segment networks: Towards good practices for deep action recognition. In: Proceedings of the European Conference on Computer Vision. pp. 20--36 (2016)

\bibitem{can}
Wang, Q., Liu, L., Jing, C., Chen, H., Liang, G., Wang, P., Shen, C.: Learning conditional attributes for compositional zero-shot learning. In: Proceedings of the IEEE/CVF Conference on Computer Vision and Pattern Recognition. pp. 11197--11206 (2023)

\bibitem{s3d}
Xie, S., Sun, C., Huang, J., Tu, Z., Murphy, K.: Rethinking spatiotemporal feature learning: Speed-accuracy trade-offs in video classification. In: Proceedings of the European Conference on Computer Vision. pp. 305--321 (2018)

\bibitem{xu2023learning}
Xu, T., Zhu, X.F., Wu, X.J.: Learning spatio-temporal discriminative model for affine subspace based visual object tracking. Visual Intelligence  \textbf{1}(1), ~4 (2023)

\bibitem{zsar2017transductive}
Xu, X., Hospedales, T., Gong, S.: Transductive zero-shot action recognition by word-vector embedding. International Journal of Computer Vision  \textbf{123},  309--333 (2017)

\bibitem{yan2022look}
Yan, R., Huang, P., Shu, X., Zhang, J., Pan, Y., Tang, J.: Look less think more: Rethinking compositional action recognition. In: Proceedings of the ACM International Conference on Multimedia. pp. 3666--3675 (2022)

\bibitem{yan2023progressive}
Yan, R., Xie, L., Shu, X., Zhang, L., Tang, J.: Progressive instance-aware feature learning for compositional action recognition. IEEE Transactions on Pattern Analysis and Machine Intelligence  (2023)

\bibitem{aim}
Yang, T., Zhu, Y., Xie, Y., Zhang, A., Chen, C., Li, M.: Aim: Adapting image models for efficient video action recognition. arXiv preprint arXiv:2302.03024  (2023)

\bibitem{yun2019cutmix}
Yun, S., Han, D., Oh, S.J., Chun, S., Choe, J., Yoo, Y.: Cutmix: Regularization strategy to train strong classifiers with localizable features. In: Proceedings of the IEEE/CVF International Conference on Computer Vision. pp. 6023--6032 (2019)

\bibitem{sampling_bias_icml}
Zadrozny, B.: Learning and evaluating classifiers under sample selection bias. In: Proceedings of the International Conference on Machine Learning. p.~114 (2004)

\bibitem{zsar2017zero}
Zellers, R., Choi, Y.: Zero-shot activity recognition with verb attribute induction. In: Proceedings of the Conference on Empirical Methods in Natural Language Processing. pp. 946--958 (2017)

\bibitem{coop}
Zhou, K., Yang, J., Loy, C.C., Liu, Z.: Learning to prompt for vision-language models. International Journal of Computer Vision  \textbf{130}(9),  2337--2348 (2022)

\end{thebibliography}

\end{document}